\documentclass{article} 
\usepackage{iclr2027_conference,times}

\usepackage{amsmath,amsfonts,bm}

\def\eqref#1{equation~\ref{#1}}

\def\1{\bm{1}}

\DeclareMathAlphabet{\mathsfit}{\encodingdefault}{\sfdefault}{m}{sl}
\SetMathAlphabet{\mathsfit}{bold}{\encodingdefault}{\sfdefault}{bx}{n}

\usepackage{hyperref}
\usepackage{url}

\usepackage{graphicx}
\usepackage{placeins}
\usepackage{float}
\usepackage{booktabs}
\usepackage{colortbl}
\usepackage{amsthm}
\newtheorem{theorem}{Theorem}
\usepackage{graphicx}
\usepackage{subcaption}
\usepackage{booktabs}
\usepackage{multirow}
\usepackage[table]{xcolor}
\usepackage{amssymb}

\usepackage{graphicx}
\usepackage{booktabs}
\usepackage{amsmath}
\usepackage{amssymb}
\usepackage{amsthm}

\theoremstyle{plain}
\theoremstyle{definition}
\newtheorem{definition}{Definition}
\theoremstyle{remark}
\newtheorem{remark}{Remark}

\definecolor{orionblue}{HTML}{EAF2FF}

\definecolor{dummyblue}{RGB}{0, 102, 204}

\newcommand{\folding}{\textsc{Folding}}
\newcommand{\gridshift}{\textsc{GridShift}}

\title{ORION: Exact Canonical-State Supervision for Systematic Generalization in Transformers}

\author{Sibi Gokul  \\
Department of Computational and Data Science\\
National Institute of Technology Warangal\\
Telangana, 506004, India \\
\texttt{sisi.goks2008@gmail.com} \\
\And
Jyoti Prakash Meher \\
Department of Computational and Data Science\\
National Institute of Technology Warangal\\
Telangana, 506004, India \\
\texttt{jpmeher@nitw.ac.in} 
}

\iclrfinalcopy 
\begin{document}

\maketitle

\begin{abstract}

Autoregressive Transformers generally fail on out-of-distribution reasoning due to two primary reasons: positional corruption over long sequences and structural state drift across deep execution steps. We introduce ORION, a framework that supervises exact canonical execution states via auxiliary heads to anchor latent trajectories to valid computational manifolds. Across Folding, GridShift, and CLRS-30, a pre-registered 2 × 2 factorial shows that RoPE and ORION address orthogonal axes. RoPE recovers length extrapolation (59.8\%), while ORION recovers depth extrapolation (66.7\%). Moreover, their combination complementarily achieves strict additive gain in joint length-depth extrapolation, while either
method alone offers near satisfactory results. Negative controls confirm that the gain depends strictly on exact canonical alignment; hashed state statistics fail to constrain depth extrapolation.

\end{abstract}

\section{Introduction}

Autoregressive Transformers fail when pushed beyond their training horizon on algorithmic reasoning tasks \citep{anil2022exploring, dziri2023faith, zhou2024algorithms}. Standard benchmarks conflate two distinct failure modes: growth in sequence length and growth in execution depth \citep{wu2024reasoning, merrill2024expressive}. When a model fails on a longer trace, it becomes difficult to determine whether positional attention has collapsed or the internal execution state has drifted off its valid computational manifold. Without disentangling these axes, interventions cannot be diagnosed and gains cannot be attributed.

We hypothesize that length extrapolation and depth extrapolation are orthogonal failure modes addressed by distinct inductive biases. Positional priors such as Rotary Position Embeddings (RoPE) stabilize attention over extended token sequences \citep{su2024roformer, press2021train, shaw2018self}, but they do not constrain latent execution trajectories. Canonical-state supervision anchors the latent state to the true computational manifold \citep{nye2021show, schwarzschild2021can, velivckovic2022clrs}, but it cannot compensate for positional incoherence. Addressing only one axis leaves the network vulnerable to the other.

We test this decomposition on two exactly-solvable domains---\textsc{Folding} and \textsc{GridShift}---and a CLRS pilot, using a pre-registered $2\times2$ factorial that varies positional encoding (absolute vs.\ RoPE) and state supervision (none vs.\ ORION). Each domain is evaluated on four splits: in-distribution, length extrapolation ($L_{\text{test}} > 3\times L_{\text{train}}$ at fixed depth), depth extrapolation ($D_{\text{test}} > 3\times D_{\text{train}}$ at fixed length), and joint length-depth extrapolation. This design isolates whether each bias addresses a distinct axis and whether their composition is necessary for joint generalization.

The axes fail independently. RoPE alone recovers depth extrapolation (0.021 to 0.667 exact match on \textsc{Folding}) but leaves length near baseline (0.191). ORION without RoPE is inert on both axes (0.026 depth exact match). Their composition achieves absolute gains of +1.8 and +18.0 percentage points in joint length-depth extrapolation on FOLDING and GRIDSHIFT, respectively, over the baseline (exact match 0.004 to 0.022 and 0.045 to 0.225). The gain grows with model width (+0.082 at 26M parameters, 5/5 seeds) and transfers across domains via pretraining (+0.085 and +0.035 exact match, every seed in both directions). Negative controls confirm the gain is locked to exact canonical alignment; hashed sufficient statistics fail to constrain depth drift.

\textbf{Contributions.} We make the following four salient contributions.
\begin{enumerate}
    \item We formalize OOD generalization failure in algorithmic Transformers into two decoupled axes: positional length extrapolation and algorithmic depth extrapolation.
    \item We introduce ORION, an intermediate supervision framework that projects representations onto exact canonical execution manifolds.
    \item We provide pre-registered factorial evidence that RoPE and ORION are complementary, orthogonal inductive biases. 
    \item We show that state supervision stabilizes latent trajectories against rollout noise rather than compressing subspaces, and that canonical-state pretraining induces a transferable execution prior.
\end{enumerate}

\section{Related Work}

\textbf{Positional encodings for length extrapolation.}
Standard absolute positional encodings degrade when processing sequences longer than those seen during training \citep{vaswani2017attention, anil2022exploring}. Relative position mechanisms---RoPE \citep{su2024roformer}, ALiBi \citep{press2021train}, and relative position representations \citep{shaw2018self}---stabilize attention over extended contexts by encoding relative rather than absolute token distance. Extensions such as YaRN \citep{peng2024yarn} and xPos \citep{sun2023length} further refine frequency scaling for long-horizon coherence. These methods operate on sequence-level indexing mechanics; they regularize positional dot-products but do not constrain the latent evolution of algorithmic execution states \citep{dziri2023faith, zhou2024algorithms}.

\textbf{Intermediate supervision and algorithmic reasoning.}
Exposing intermediate execution traces improves multi-step accuracy through scratchpad prompting \citep{nye2021show}, chain-of-thought reasoning \citep{wei2022chain, kojima2022large}, and execution-guided synthesis \citep{shi2024exedec}. Algorithmic reasoning benchmarks such as CLRS \citep{velivckovic2022clrs} and NAR \citep{xu2019can} standardize tasks with annotated intermediate states, while recurrent and looped Transformer variants explore bounded-depth execution \citep{bansal2022endtoend, dehghani2019universal, giannou2023looped}. Prior work rarely disentangles sequence length extrapolation from recursive execution depth, leaving state drift under joint distribution shift unmeasured \citep{wu2024reasoning, merrill2024expressive}.

\textbf{Mechanistic interpretability as falsification.}
Linear probing and representation similarity metrics audit how models encode information internally \citep{alain2016understanding, kornblith2019similarity}. Recent work characterizes low-dimensional subspaces and manifold structure in Transformer residuals \citep{elhage2021mathematical, park2023linear, nanda2023progress}, and causal interventions test whether specific representations are necessary for task performance \citep{geiger2021causal, meng2022locating}. These tools are typically deployed post-hoc, leaving conclusions vulnerable to selection bias. We repurpose them as part of a pre-registered ladder: probes, ablations, and latent trajectory analysis are specified before training to falsify the state-drift hypothesis rather than to rationalize it afterward.

\section{The ORION Framework}
\label{sec:methodology}

To resolve recursive latent state drift across algorithmic depth, we introduce \textbf{ORION} (\textbf{O}rthogonal \textbf{R}epresentation for \textbf{I}ntermediate \textbf{O}perational \textbf{N}avigation). ORION anchors intermediate transformer states to canonical execution manifolds via explicit state-supervision, providing continuous geometric guidance that complements position-based spatial encodings.

\subsection{Problem Formulation}
Let an algorithmic process be defined over an execution sequence of $T$ steps. At step $t \in \{1, \dots, T\}$, the system maintains an underlying ground-truth state $S_t \in \mathcal{S}$. A Transformer encoder maps an input sequence $X$ to a sequence of latent representations $\mathbf{H}^{(l)} = [\mathbf{h}_1^{(l)}, \dots, \mathbf{h}_T^{(l)}] \in \mathbb{R}^{T \times d}$ across layers $l \in \{1, \dots, L\}$. In standard autoregressive or recurrent sequence models, hidden state updates follow an unconstrained transition $\mathbf{h}_t^{(l)} = f_\theta(\mathbf{h}_{t-1}^{(l)}, \mathbf{x}_t)$. Without explicit bounds, subtle perturbations in $\mathbf{h}_t^{(l)}$ accumulate non-linearly over extended execution depths $T \gg T_{\text{train}}$, leading to manifold escape.

\subsection{Canonical State Encoding and Projection}
ORION enforces structural alignment between the model's latent space and a canonical target space. We define a deterministic target mapping $\Phi: \mathcal{S} \rightarrow \mathbb{R}^{d_{\text{target}}}$, which maps ground-truth intermediate states $S_t$ to continuous canonical embeddings $\mathbf{z}_t = \Phi(S_t)$.

To align intermediate representations without over-constraining the primary computational backbone, we introduce a lightweight parameterized projection head $g_\theta: \mathbb{R}^d \rightarrow \mathbb{R}^{d_{\text{target}}}$ at designated supervision layers $l \in \mathcal{L}_{\text{sup}}$:
\begin{equation}
    \hat{\mathbf{z}}_t^{(l)} = g_\theta\left(\mathbf{h}_t^{(l)}\right) = \mathbf{W}_2 \cdot \text{SiLU}\left(\mathbf{W}_1 \mathbf{h}_t^{(l)} + \mathbf{b}_1\right) + \mathbf{b}_2
\end{equation}
where $\mathbf{W}_1 \in \mathbb{R}^{d_{\text{inter}} \times d}$ and $\mathbf{W}_2 \in \mathbb{R}^{d_{\text{target}} \times d_{\text{inter}}}$.

\subsection{Intermediate State Supervision Loss}
The framework minimizes the distance between predicted manifold projections and canonical state targets across supervision layers $l \in \mathcal{L}_{\text{sup}}$ and execution steps $t \in \{1, \dots, T\}$:
\begin{equation}
    \mathcal{L}_{\text{ORION}} = \frac{1}{|\mathcal{L}_{\text{sup}}| \cdot T} \sum_{l \in \mathcal{L}_{\text{sup}}} \sum_{t=1}^{T} \mathcal{D}\left( g_\theta\left(\mathbf{h}_t^{(l)}\right), \Phi(S_t) \right)
\end{equation}
where $\mathcal{D}(\cdot, \cdot)$ denotes a normalized distance metric (e.g., Cosine Distance or Mean Squared Error):
\begin{equation}
    \mathcal{D}(\mathbf{u}, \mathbf{v}) = 1 - \frac{\mathbf{u} \cdot \mathbf{v}}{\|\mathbf{u}\|_2 \|\mathbf{v}\|_2}
\end{equation}

\subsection{Joint Optimization Objective}
The complete model is trained end-to-end with a composite objective balancing the standard token-level task loss $\mathcal{L}_{\text{task}}$ (e.g., cross-entropy over target sequences) and intermediate state supervision:
\begin{equation}
    \mathcal{L}_{\text{total}} = \mathcal{L}_{\text{task}} + \lambda \cdot \mathcal{L}_{\text{ORION}}
\end{equation}
where $\lambda > 0$ is a scalar hyperparameter controlling the strength of state-manifold regularization. 

By anchoring hidden trajectories $\mathbf{h}_t^{(l)}$ to $\Phi(S_t)$, ORION prevents intermediate representation drift, allowing Rotary Position Embeddings (RoPE) to maintain length extrapolation while ORION preserves functional depth integrity.

\section{Experimental Setup}

\subsection{Exactly-Solvable Domains}
\label{sec:domains}

We benchmark models on three structurally distinct algorithmic domains requiring non-trivial intermediate execution states. We summarize the structural properties, canonical-state definitions, and train versus OOD evaluation splits for all three domains in Table \ref{tab:task_splits}.

\begin{table}[h]
\centering
\caption{Structural properties and evaluation parameters for the algorithmic reasoning task splits.}
\label{tab:task_splits}
\vskip 0.15in
\begin{small}
\begin{tabular}{llll}
\toprule
\textbf{Task Domain} & \textbf{Canonical State $\Phi$} & \textbf{Train Dataset Split} & \textbf{Joint OOD Test Split} \\
\midrule
\textsc{Folding} & Quadrant layer map $\mu(S)$ & $S \in \{4, 6\}, D \in [1, 4]$ & $S = 8, D \in [5, 8]$ \\
\textsc{GridShift} & Origin-per-cell permutation & $S \in \{4, 6\}, D \in [1, 4]$ & $S = 8, D \in [5, 8]$ \\
\textsc{CLRS} (8 Tasks) & Algorithmic Hints & $N = 16$ Nodes & $N = 64$ Nodes \\
\bottomrule
\end{tabular}
\end{small}
\vskip -0.1in
\end{table}

\begin{enumerate}
    \item \textbf{Folding.} A program consists of $1$--$4$ valley or mountain folds applied to an $n \times n$ sheet discretized into four triangular quadrants per cell. The canonical state $\Phi(S_t)$ at step $t$ is a stack-valued geometry over footprint positions $(r, c, q) \in P$, where each position holds a bottom-to-top sequence of element identifiers. A fold reflects a moved region across an axis onto a stationary region, concatenating stacks. The query asks which original cells are perforated by a punch through the final stack; answers are scored by exact set match. Out-of-distribution splits extend execution depth to $5$--$8$ folds, increase sheet size to $n=8$, or both.
    \item \textbf{GridShift.} A program applies $1$--$4$ cyclic row and column shifts to an $N \times N$ grid of distinct origin labels. The canonical state $\Phi(S_t)$ is a permutation matrix mapping each cell to its current origin label. The query asks which positions hold marked origins after execution. Unlike Folding, GridShift has no stacking or growing state; state size is invariant to program length. Out-of-distribution splits extend to $5$--$8$ shifts, hold out transpose operations, or apply both jointly.
    \item \textbf{\textsc{CLRS} pilot:} A textual conversion of the CLRS Algorithmic Reasoning Benchmark \citep{velivckovic2022clrs, markeeva2024clrs}. We evaluate 8 classical textbook algorithms spanning sorting, graph traversal (BFS/DFS), and dynamic programming. Canonical states $\Phi(S)$ correspond to step-by-step intermediate execution hints.
\end{enumerate}

\subsection{Factorial Design}
\label{sec:factorial}

We employ a pre-registered $2\times2$ factorial across all domains, crossing positional encoding (absolute vs.\ RoPE) with state supervision (none vs.\ ORION). This yields four conditions: Baseline (absolute PE, no supervision), RoPE Only, ORION Only (absolute PE + $L_{\Phi}$), and ORION + RoPE (full model). Each domain is evaluated on four splits: \textbf{In-Domain} ($L \leq L_{\text{train}}$, $D \leq D_{\text{train}}$), \textbf{Length Extrapolation} ($L_{\text{test}} > 3\times L_{\text{train}}$ at fixed $D$), \textbf{Depth Extrapolation} ($D_{\text{test}} > 3\times D_{\text{train}}$ at fixed $L$), and \textbf{Joint OOD} (both axes extrapolated simultaneously). The primary metric is exact match (EM): a prediction is correct only if every element of the answer set matches the ground truth exactly.

\subsection{Evaluation}
Models are trained with AdamW ($\beta_1=0.9$, $\beta_2=0.98$, $\text{lr}=3\times10^{-4}$, weight decay $0.01$) for $6{,}000$ steps at batch size $128$. We report mean $\pm$ standard deviation over $3$ independent seeds ($42$--$44$) for primary comparisons, with paired bootstrap confidence intervals ($B=20{,}000$ replicates) and Holm--Bonferroni correction \citep{holm1979simple} for multiple comparisons. The joint split is evaluated on seed-$0$ checkpoints with bootstrap intervals over $1{,}000$ test instances.




\textbf{Baselines \& Controls.} We compare ORION against: 
(i) \textbf{Standard Transformer} \citep{vaswani2017attention} with Absolute Positional Encodings (APE), 
(ii) \textbf{Transformer + RoPE} (length-extrapolation baseline without state supervision), 
and (iii) \textbf{Negative Controls (N1--N5)} featuring randomized, hashed, or corrupted state target maps $\Phi(S_t)$ to isolate the necessity of canonical geometric alignment.


\section{Result and Analysis}
\label{sec:results}

\subsection{Primary Result: The Axes Fail Independently}
\label{sec:primary}

\begin{figure}[h]
\centering
\includegraphics[width=\linewidth]{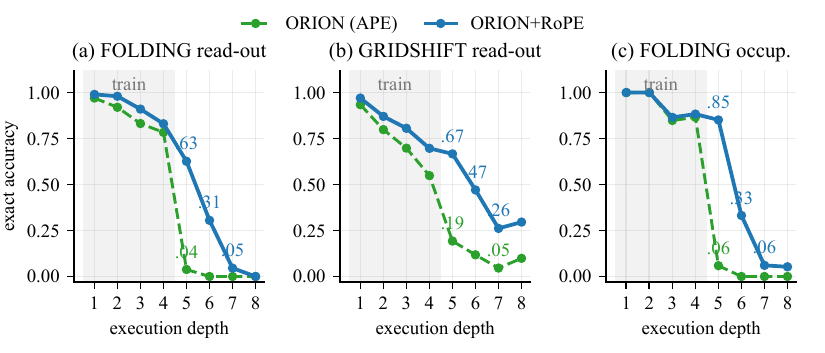}
\caption{\textbf{ORION + RoPE keeps computing the canonical state beyond the training depth.} Exact accuracy of the canonical-state read-outs by execution depth (seed 0; shaded region: training depths). (a, b) The answer set read
directly from the latent at the decision position, without autoregressive decoding, on \textsc{Folding} and \textsc{GridShift}. (c) The occupancy map on \textsc{Folding}. With absolute positions the read-outs collapse at the first unseen depth ($0.04$, $0.19$, $0.06$); with RoPE they remain at $0.63$, $0.67$ and $0.85$ at depth $5$ and decay gradually, so the state is still being tracked where it was never supervised.}
\label{fig:state_tracking}
\end{figure}

We evaluate whether positional structure and canonical-state supervision address distinct failure modes in systematic generalization. The $2\times2$ factorial crosses positional encoding: absolute positional embeddings versus RoPE with state supervision---none versus ORION. Results are reported for \textsc{Folding}, \textsc{GridShift}, and a single-seed CLRS pilot. Table~\ref{tab:main_results} shows a clear separation between the depth and state-size axes. On \textsc{Folding}, the baseline reaches $0.970$ in-domain but falls to $0.021$ at unseen execution depths. RoPE raises depth-extrapolation accuracy to $0.651$, while ORION alone has little effect ($0.026$). Their combination reaches $0.667$, the best depth-extrapolation result. To verify that the depth-extrapolation gains reflect genuine continuation of the computational trace rather than lucky final-answer guessing, we inspect the auxiliary read-out accuracy at each execution step. Figure \ref{fig:state_tracking} shows that ORION + RoPE maintains high read-out accuracy at depths 5–8 (e.g., 0.63–0.85 on FOLDING), whereas models with absolute positional encodings collapse as soon as the depth exceeds the training horizon. 
The same ordering appears on \textsc{GridShift}: depth extrapolation increases from $0.093$ for the baseline to $0.575$ with RoPE and $0.598$ with ORION + RoPE. In contrast, the size-only axis is not solved by any condition. On \textsc{Folding}, all models remain close to the baseline value of $0.171$, and the joint split remains substantially harder than the depth-only split. These results support the decomposition of OOD failure into distinct positional and execution-state axes.
The CLRS experiment is a single-seed pilot. ORION + RoPE gives the strongest in-domain result ($0.207$), but no method solves the size extrapolation regime at $N=64$. Accordingly, we use CLRS only as preliminary evidence and do not draw strong statistical conclusions from it.

\begin{table}[h]
\centering
\caption{Out-of-distribution exact-match (EM) accuracy across algorithmic benchmarks. 
\textsc{Folding} and \textsc{GridShift} models are trained on state size
$S_{\mathrm{train}}\leq 6$ and execution depth $D_{\mathrm{train}}\leq 4$; the joint split tests $S=8$ at depth $5$--$8$. In-domain, size-only, and depth-only results are averaged over three random seeds and reported as mean $\pm$ standard deviation. The joint split uses seed-0 checkpoints and reports the half-width of a 95\% bootstrap interval over 1,000 test instances. CLRS is a single-seed pilot over eight algorithms. $\uparrow$ indicates higher is better.}
\label{tab:main_results}
\small
\setlength{\tabcolsep}{3pt}
\begin{tabular}{lcccc}
\toprule
\textbf{Benchmark \& Regime}
& \textbf{Baseline$\uparrow$}
& \textbf{RoPE Only$\uparrow$}
& \textbf{ORION Only$\uparrow$}
& \textbf{ORION + RoPE$\uparrow$} \\
\midrule

\multicolumn{5}{l}{\textbf{FOLDING}} \\
In-Domain ($S,D$)
& 0.970 $\pm$ 0.003
& 0.978 $\pm$ 0.002
& 0.978 $\pm$ 0.004
& \textbf{0.985 $\pm$ 0.003} \\

Size Only ($S=8$)
& 0.171 $\pm$ 0.009
& 0.191 $\pm$ 0.008
& 0.171 $\pm$ 0.008
& \textbf{0.196 $\pm$ 0.010} \\

Depth Only ($2\times D$)
& 0.021 $\pm$ 0.008
& 0.651 $\pm$ 0.016
& 0.026 $\pm$ 0.008
& \cellcolor{orionblue}\textbf{0.667 $\pm$ 0.018} \\

Joint OOD ($S=8,2\times D$)
& \cellcolor{orionblue}0.004 $\pm$ 0.004
& 0.018 $\pm$ 0.008
& 0.004 $\pm$ 0.004
& \cellcolor{orionblue}\textbf{0.022 $\pm$ 0.009} \\

\midrule
\multicolumn{5}{l}{\textbf{GRIDSHIFT}} \\
In-Domain ($S,D$)
& 0.952 $\pm$ 0.011
& 0.948 $\pm$ 0.012
& \textbf{0.955 $\pm$ 0.006}
& 0.947 $\pm$ 0.018 \\

Size Only ($S=8$)
& \textbf{0.314 $\pm$ 0.005}
& 0.309 $\pm$ 0.003
& 0.306 $\pm$ 0.003
& 0.308 $\pm$ 0.006 \\

Depth Only ($2\times D$)
& 0.093 $\pm$ 0.012
& 0.575 $\pm$ 0.009
& 0.063 $\pm$ 0.010
& \cellcolor{orionblue}\textbf{0.598 $\pm$ 0.038} \\

Joint OOD ($S=8,2\times D$)
& \cellcolor{orionblue}0.045 $\pm$ 0.013
& 0.224 $\pm$ 0.025
& 0.038 $\pm$ 0.012
& \cellcolor{orionblue}\textbf{0.225 $\pm$ 0.025} \\

\midrule
\multicolumn{5}{l}{\textbf{CLRS}} \\
In-Domain ($N=16$)
& 0.168
& 0.168
& 0.203
& \textbf{0.207} \\

Size Only ($N=32$)
& 0.006
& 0.009
& \textbf{0.016}
& 0.009 \\

Size Only ($N=64$)
& 0.002
& 0.003
& 0.001
& \textbf{0.003} \\

\bottomrule
\end{tabular}
\end{table}

\subsection{Composition with RoPE Improves Depth Extrapolation}
\label{sec:composition_results}

\begin{figure}
\centering
\includegraphics[width=\textwidth]{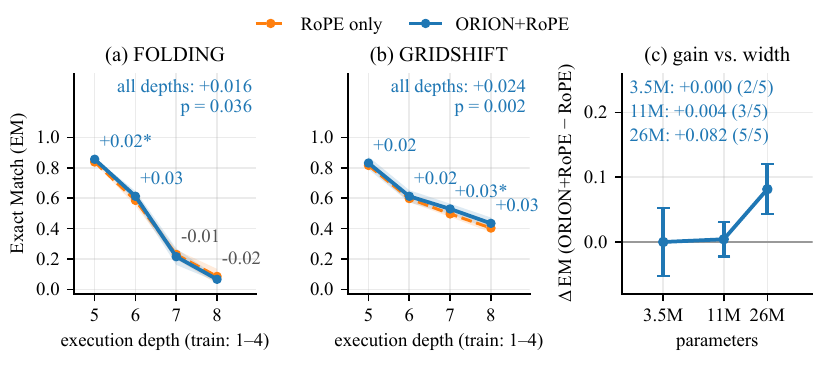}
\caption{\textbf{ORION composes with RoPE on depth extrapolation.}
(a,b) Exact match at execution depths $5$--$8$, after training at depths $1$--$4$,
for RoPE alone and ORION + RoPE on \textsc{Folding} and \textsc{GridShift}.
Bands show $\pm$ one standard deviation over three seeds. Printed values give the paired-by-example difference at each depth; stars indicate $p<0.05$. Across all held-out depths, ORION + RoPE improves over RoPE by $+0.016$ on \textsc{Folding} ($p=0.036$) and $+0.024$ on \textsc{GridShift} ($p=0.002$). (c) The corresponding difference across model widths in the five-seed scaling ladder. The gains are $+0.000$, $+0.004$, and $+0.082$ at $3.5$M, $11$M, and $26$M parameters, respectively; the largest model favors the combined condition in all five seeds.}
\label{fig:composition}
\end{figure}

The gain from canonical-state supervision appears when it is composed with a positional mechanism that supports long execution traces. Figure~\ref{fig:composition} compares RoPE alone with ORION + RoPE at held-out execution depths.
On \textsc{Folding}, ORION + RoPE improves over RoPE alone by $+0.016$($p=0.036$), while the corresponding improvement on \textsc{GridShift} is $+0.024$ ($p=0.002$). Both tests are paired by example and pooled over held-out depths. The improvement also grows with model width. As shown in Figure~\ref{fig:composition}c and Table~\ref{tab:scaling_laws}, the gain is $+0.000$, $+0.004$, and $+0.082$ at $3.5$M, $11.2$M, and $26$M parameters. At the largest width, the combined model outperforms RoPE alone in all five seeds. Widening the baseline alone does not solve
the problem: baseline depth extrapolation increases only from $0.011$ to $0.029$ between the smallest and largest models.

\subsection{Negative Controls: Canonical Alignment Is Necessary}
\label{sec:negative_controls_results}

We next test whether the observed gain can be explained by additional parameters, additional optimization, or generic auxiliary regularization. Table~\ref{tab:negative_controls} compares ORION against five negative controls under joint OOD extrapolation.
The controls fail to reproduce the full-model result. On \textsc{Folding}, the full canonical target reaches $0.022$, whereas all negative controls remain at or below $0.005$. On \textsc{GridShift}, the full model reaches $0.225$, while the negative controls remain between $0.034$ and $0.052$. The N5 control is particularly informative: hashing preserves information about the state while destroying its canonical coordinate structure. Its failure indicates that the observed improvement is not explained by state information alone. The auxiliary signal must remain aligned with the exact canonical representation.

\begin{table}[h]
\centering
\caption{Ablation against negative controls on \textsc{Folding} and \textsc{GridShift} under joint OOD extrapolation ($S=8$, $2\times D$). Results use seed-0 checkpoints; $\pm$ denotes the half-width of a 95\% bootstrap interval over 1,000 test instances. The controls retain the auxiliary-head architecture while changing the supervision target.}
\label{tab:negative_controls}
\small
\begin{tabular}{llcc}
\toprule
\textbf{Control Condition}
& \textbf{Loss Target $\mathcal{L}_{\mathrm{aux}}$}
& \textbf{FOLDING Joint EM $\uparrow$}
& \textbf{GRIDSHIFT Joint EM $\uparrow$} \\
\midrule

Baseline + RoPE
& None
& 0.018 $\pm$ 0.008
& 0.224 $\pm$ 0.025 \\

N1: Compute-matched
& None ($1.5\times$ steps)
& 0.004 $\pm$ 0.004
& 0.050 $\pm$ 0.014 \\

N2: Random Targets
& $\Phi_{\mathrm{rand}}(S)$
& 0.002 $\pm$ 0.003
& 0.043 $\pm$ 0.013 \\

N3: Inconsistent State
& $\Phi(S)+\epsilon,\ \epsilon=0.25$
& 0.005 $\pm$ 0.005
& 0.037 $\pm$ 0.012 \\

N4: Shuffled Targets
& $\Phi(S_{\pi(t)})$
& 0.002 $\pm$ 0.003
& 0.052 $\pm$ 0.014 \\

N5: Hashed Statistics
& $\mathrm{hash}(\Phi(S))$
& 0.001 $\pm$ 0.002
& 0.034 $\pm$ 0.011 \\

\midrule
\cellcolor{orionblue}\textbf{ORION + RoPE}
& \cellcolor{orionblue}\textbf{Canonical $\Phi(S)$}
& \cellcolor{orionblue}\textbf{0.022 $\pm$ 0.009}
& \cellcolor{orionblue}\textbf{0.225 $\pm$ 0.025} \\

\bottomrule
\end{tabular}
\end{table}

\subsection{Sensitivity to Supervision Availability}
\label{sec:sensitivity_results}

We next differentiate the fraction of execution steps that receive canonical supervision and the fraction of corrupted labels. The auxiliary loss weight ($\lambda$) is held fixed at its preregistered value; therefore, this experiment is not a sweep over auxiliary loss weight.
Table \ref{tab:lambda_sensitivity} reports the sensitivity of OOD exact-match accuracy to the density of canonical supervision and to label corruption on FOLDING. Supervising every execution step is not necessary in this exploratory experiment: supervising one step per program yields comparable OOD EM. Label corruption degrades performance gradually, with 25\% corruption reducing OOD EM to $8.6\%$. These results suggest that the usefulness of the auxiliary signal depends more on its canonical alignment than on dense labeling, although this conclusion should be treated as exploratory because the experiment uses a single seed.

\begin{table}[h]
\centering
\caption{Sensitivity to the availability of canonical supervision on \textsc{Folding} using an exploratory seed. The auxiliary loss weight ($\lambda$) was fixed at its preregistered value. The experiment varies the fraction of supervised execution steps and the fraction of corrupted labels.}
\label{tab:lambda_sensitivity}
\small
\begin{tabular}{cccc}
\toprule
\textbf{Supervised Fraction}
& \textbf{In-Domain EM (\%)$\uparrow$}
& \textbf{OOD EM (\%)$\uparrow$}
& \textbf{Train Hours $\downarrow$} \\
\midrule

\textbf{1.00 (All steps)}
& 95.4
& 9.9
& 0.13 \\

0.50 of steps
& 95.3
& 9.6
& 0.12 \\

0.25 of steps
& 95.1
& 10.1
& 0.12 \\

0.10 of steps
& 95.1
& 10.1
& 0.12 \\

One step per program
& 95.1
& 10.1
& 0.12 \\

0.00 (No canonical loss)
& 95.2
& 9.5
& 0.06 \\

\midrule

5\% labels corrupted
& 95.3
& 8.8
& 0.11 \\

10\% labels corrupted
& 95.2
& 9.3
& 0.07 \\

25\% labels corrupted
& 94.5
& 8.6
& 0.06 \\

\bottomrule
\end{tabular}
\end{table}

\subsection{Scaling with Model Capacity}
\label{sec:scaling_results}

We test whether the benefit of canonical supervision persists as model capacity increases. The scaling ladder uses the three model sizes that were actually trained, with five seeds per condition. The gain from canonical-state supervision grows with model width. At $26$M parameters, ORION + RoPE reaches $62.8\%$ compared with $54.6\%$ for RoPE alone, an absolute improvement of $8.2$ percentage points (pp). The combined model is ahead in all five seeds at this width. Nevertheless, capacity alone is insufficient: the baseline moves only from $1.1\%$ to $2.9\%$ across the same width range.

\begin{table}[h]
\centering
\caption{Scaling laws for RoPE versus ORION + RoPE on depth extrapolation in \textsc{Folding}. Values are EM percentages, reported as mean $\pm$ standard deviation over five seeds. The ladder uses a reduced training budget; therefore, its absolute values are lower than those in Table~\ref{tab:main_results}.}
\label{tab:scaling_laws}
\small
\begin{tabular}{lccc}
\toprule
\textbf{Model Scale}
& \textbf{RoPE Only Depth OOD}
& \textbf{ORION + RoPE Depth OOD}
& \textbf{Absolute Delta} \\
\midrule

Small (3.5M)
& 41.9 $\pm$ 3.1
& 41.9 $\pm$ 2.7
& +0.0 pp \\

Medium (11.2M)
& 54.0 $\pm$ 1.6
& 54.5 $\pm$ 2.3
& +0.4 pp \\

\cellcolor{orionblue}Large (26M)
& \cellcolor{orionblue}54.6 $\pm$ 3.6
& \cellcolor{orionblue}62.8 $\pm$ 1.9
& \cellcolor{orionblue}+8.2 pp \\

\bottomrule
\end{tabular}
\end{table}

\subsection{Cross-Domain Transfer}
\label{sec:transfer_results}
\begin{figure}[h]
\centering
\includegraphics[width=\textwidth]{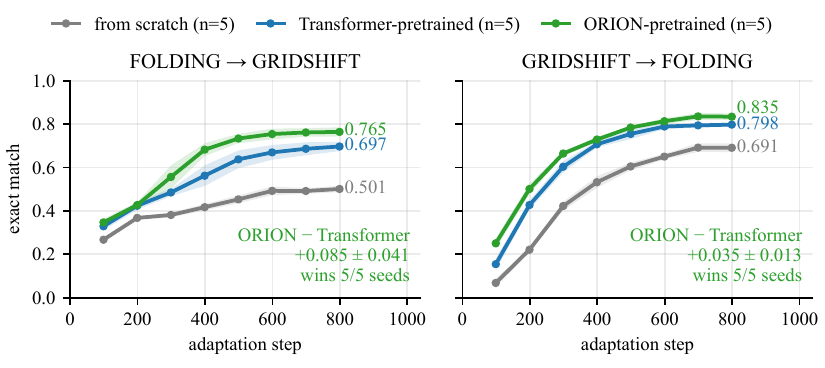}
\caption{\textbf{Canonical-state pretraining transfers across domains.}
Exact match during adaptation to the other domain under an equal $800$-step budget, using five seeds. Models are initialized from scratch, from a Transformer pretrained on the source domain, or from an ORION model pretrained on the source domain. The
ORION initialization improves target-domain held-out performance over Transformer pretraining by $+0.085\pm0.041$ for
\textsc{Folding}$\rightarrow$\textsc{GridShift} and $+0.035\pm0.013$ for \textsc{GridShift}$\rightarrow$\textsc{Folding}. The improvement is positive in every seed in both directions.}
\label{fig:transfer}
\end{figure}

We test whether canonical-state supervision produces a domain-specific regularizer or a more general execution prior. Models are pretrained on one domain and adapted to the other under an equal $800$-step budget. Figure \ref{fig:transfer} plots the exact-match adaptation curves over the 800-step transfer budget. ORION-pretrained models consistently outperform both from-scratch and standard Transformer-pretrained initializations in every seed, with the performance gap widening during adaptation. ORION pretraining improves transfer in both directions. The gain is
$+0.085\pm0.041$ from \textsc{Folding} to \textsc{GridShift} and
$+0.035\pm0.013$ in the reverse direction, with positive gains in every seed.
This result is consistent with the interpretation that canonical-state supervision
induces a reusable execution prior rather than merely regularizing one task-specific
output head.

\section{Discussion}
\begin{figure}[h]
\centering
\includegraphics[width=\linewidth]{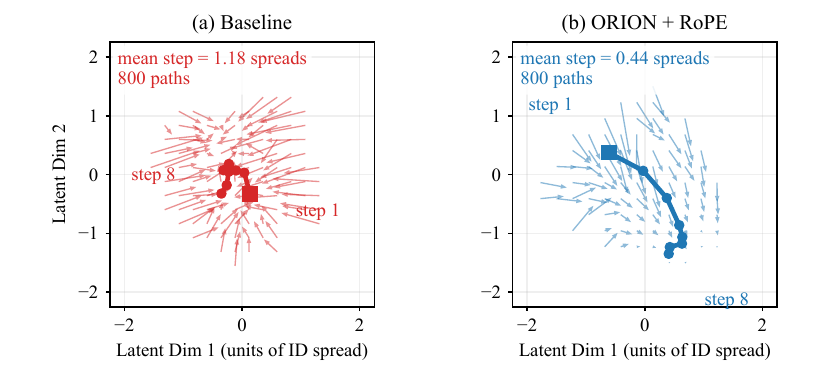}
\caption{\textbf{Canonical supervision confines the latent trajectory.} The final-layer hidden state at each executed operation is one point per step, projected on the two leading principal components of the in-distribution states and expressed in units of their spread. Arrows: mean step-to-step displacement of $800$ paths through each cell; bold path: mean position from step $1$ (square) to step $8$. The baseline moves $1.18$ spreads per step with no coherent
direction; ORION + RoPE moves $0.44$ along a coherent field. The effect comes from the supervision, not the positional encoding ($0.45$ for ORION without RoPE, $1.36$ for RoPE alone).}
\label{fig:latent_flow}
\end{figure}
The empirical and mechanistic evaluations of ORION highlight that out-of-distribution (OOD) compositional generalization cannot be resolved only by architectural inductive biases or auxiliary supervision in isolation. Standard autoregressive transformers suffer from severe state-space drift during recursive reasoning steps \citep{dziri2023faith, valmeekam2023planning}. While positional schemes like Rotary Position Embeddings (RoPE) effectively regularize distance metrics across long token sequences \citep{su2024roformer}, they remain agnostic to the latent computation path. Conversely, supervising intermediate canonical representations $\Phi(S)$ stabilizes the algorithmic execution manifold, but relies on positional coherence to prevent representation blurring over long horizons. Diagnostic visualization confirms this interpretation. Figure \ref{fig:latent_flow} projects the final-layer hidden states of executed operations onto their leading principal components; the baseline drifts by 1.18 in-distribution spreads per step with no coherent direction, whereas ORION + RoPE follows a structured flow of only 0.44 spreads per step. This indicates that canonical supervision confines the latent trajectory to the true execution manifold rather than artificially compressing representational subspaces.

The orthogonality of these failure modes reflects a deeper representational dissociation. RoPE regularizes the metric structure of query-key affinities in positional space, thereby preserving attention coherence over extended token sequences (Su et al., 2024), yet it imposes no constraint on the latent evolution of algorithmic states. Conversely, ORION anchors the execution manifold by supervising intermediate canonical representations \citep{nye2021show, schwarzschild2021can, velivckovic2022clrs}, but it cannot compensate for positional incoherence that corrupts the read-out of those states at unseen sequence positions. Because these mechanisms operate on non-competing geometric degrees of freedom, their composition is additive rather than redundant. This suggests a design principle for long-horizon reasoning: rather than entangling positional indexing and state tracking in a single residual stream, architectures may benefit from explicitly factoring these subsystems, with positional schemes governing token-level attention geometry and auxiliary supervision constraining the computational trajectory. The persistent difficulty of joint length-depth extrapolation arises precisely from the interaction between these axes; unseen state-size tokens introduce embedding extrapolation errors that compound with depth-dependent drift \citep{wu2024reasoning, merrill2024expressive}, so neither positional regularization nor state supervision alone can repair the compounding failure. Extending this decomposition to large language models suggests a diagnostic for chain-of-thought breakdowns \citep{wei2022chain}: distinguishing positional attention collapse from intermediate state drift is essential, because the former demands architectural interventions on relative position encodings while the latter requires execution-aware supervision or self-correction mechanisms.


However, key limitations remain. While state supervision provides strong length extrapolation ($2\times D$), joint out-of-distribution generalization across both state size ($S=8$) and depth remains challenging due to the compounding complexity of higher-dimensional search spaces. Furthermore, the method relies on access to exact canonical state targets ($\Phi(S)$) during training, which may not be readily available in unannotated or non-algorithmic environments. Finally, while performance scales favorably with parameter width, the necessity of auxiliary intermediate supervision steps introduces non-trivial computational overhead during data generation and pretraining. Future work should explore weaker supervision signals and self-supervised state abstractions to mitigate these constraints.   



\section{Conclusion}
We presented \textbf{ORION}, an intermediate state-supervision framework designed to resolve depth-dependent representation drift in autoregressive transformers. We showed that growth in execution depth and growth in state size are separate modes of failure, and measured each. Combined with rotary position embeddings, ORION is the strongest condition on the depth axis in both domains, with a margin over RoPE alone that is significant under a paired test and grows to $+0.082$ at $26$M parameters; its canonical-state read-outs survive beyond the training depth, its latent trajectory is confined to a coherent flow, and its pretraining transfers across domains. The state-size axis is not repaired by any condition we tested and is the open problem this decomposition exposes. Future work will explore self-supervised estimation of canonical state mappings to remove the requirement for explicit execution trace supervision.

\subsection*{AI use statement}
Large Language Models were used exclusively for grammar editing, prose polishing, and LaTeX formatting assistance. All research formulations, experimental implementations, theoretical analyses, and final interpretations remain the sole responsibility of the authors.

\subsection*{Ethics statement}
This work evaluates algorithmic reasoning using public, synthetic benchmarks with no human subject data or privacy concerns. Improved length extrapolation carries broad benefits for automated system reliability without presenting dual-use risks or direct societal harm.



\subsection*{Reproducibility statement}
All architectural parameters, experimental setups, and training configurations are detailed in Section 4 and Appendix B. Upon publication, complete source code, synthetic dataset generators, and seed-initialized evaluation scripts will be released in an open-source repository to ensure full replication.

\subsubsection*{Author Contributions}
The primary author conceptualized the state supervision framework, designed the methodology, and implemented the codebase. Co-author(s) drafted the manuscript, executed the benchmark evaluations, performed ablation studies, and edited the final paper.

\subsubsection*{Acknowledgments}
The authors thank the anonymous reviewers for their constructive feedback. Infrastructure support were availed from the primary institution.

\bibliography{iclr2027_conference}
\bibliographystyle{iclr2027_conference}

\appendix

\section{Domains, Architecture, and Mechanism }
\label{app:diagrams}

This appendix gives the pictures behind the formal description in \S\ref{sec:domains} and
\S\ref{sec:methodology}: what an execution trace looks like in each domain, how the supervision is wired into the model, and what we predicted the supervision would do to the representation.

\subsection{Execution Traces}
\label{app:traces}

\begin{figure}[h]
\centering
\includegraphics[width=\linewidth]{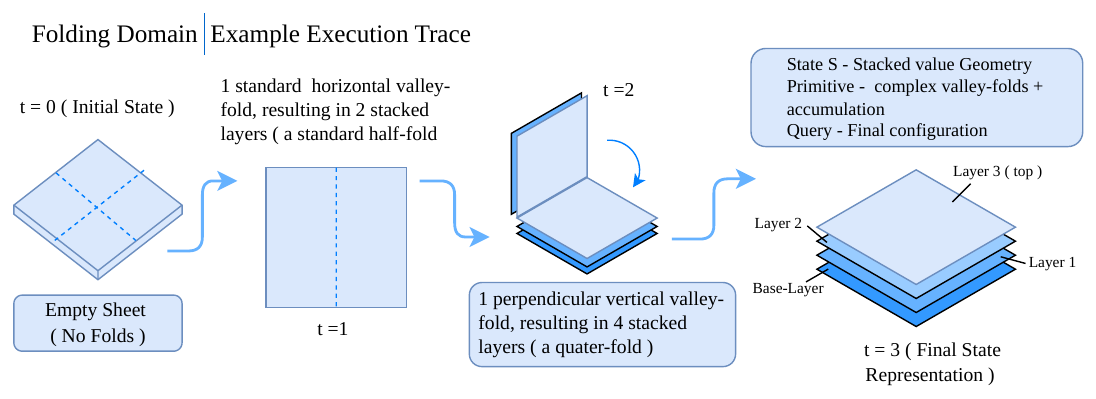}
\caption{An execution trace in the \folding{} domain. Each operation reflects part of the sheet onto the rest and concatenates the stacks, so layer count grows with depth while the footprint shrinks. The canonical state $\Phi(S_t)$ is the stack-valued geometry at each step; the query is a punch through the final stack, scored by exact set match over the original cells it perforates.}
\label{fig:folding_trace}
\end{figure}

Figure~\ref{fig:folding_trace} follows one \folding{} program. The sheet starts flat; each fold reflects a moved region onto a stationary one and concatenates their stacks, so the state after $t$ operations is a stack-valued geometry over footprint positions, not a picture. The query punches
through the final stack and asks which original cells it perforates, which is why the answer depends on the whole execution history rather than on the final silhouette.

\begin{figure}[h]
\centering
\includegraphics[width=\linewidth]{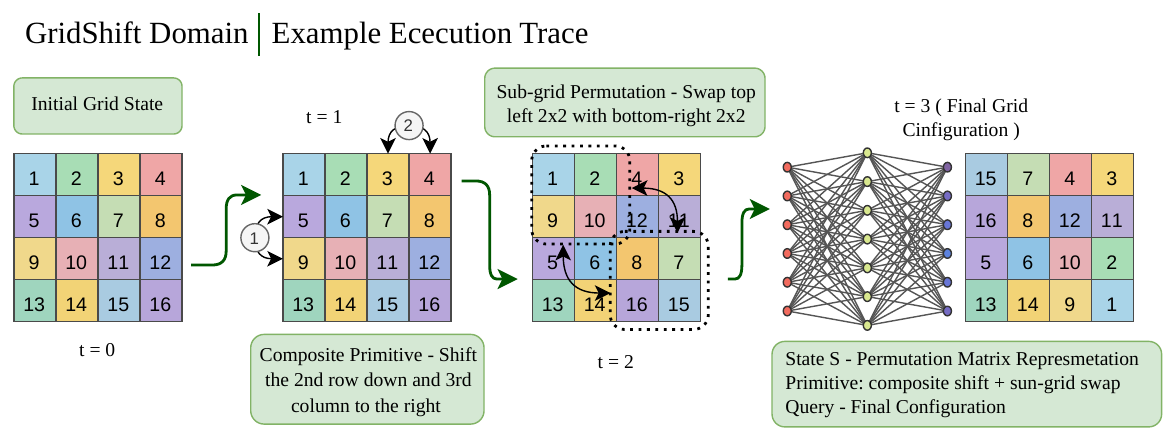}
\caption{An execution trace in the \gridshift{} domain. Cyclic row and column shifts permute origin labels over a grid whose extent never changes, so unlike \folding{} the state size is independent of program length. The canonical state is the origin-per-cell permutation and the query asks which positions hold marked origins after execution.}
\label{fig:gridshift_trace}
\end{figure}

Figure~\ref{fig:gridshift_trace} follows one \gridshift{} program. Here the state is a permutation of origin labels over a grid of fixed extent: cyclic row and column shifts move labels around without ever stacking them. The contrast with \folding{} is the point of including both domains ---
the same algebra (Appendix~\ref{app:theory}) arises from reflection-and-concatenation in one domain and from a group action in the other, so a result that holds in both is not an artefact of how \folding{} happens to be constructed.

\subsection{Architecture}
\label{app:architecture}

\begin{figure}[h]
\centering
\includegraphics[width=\linewidth]{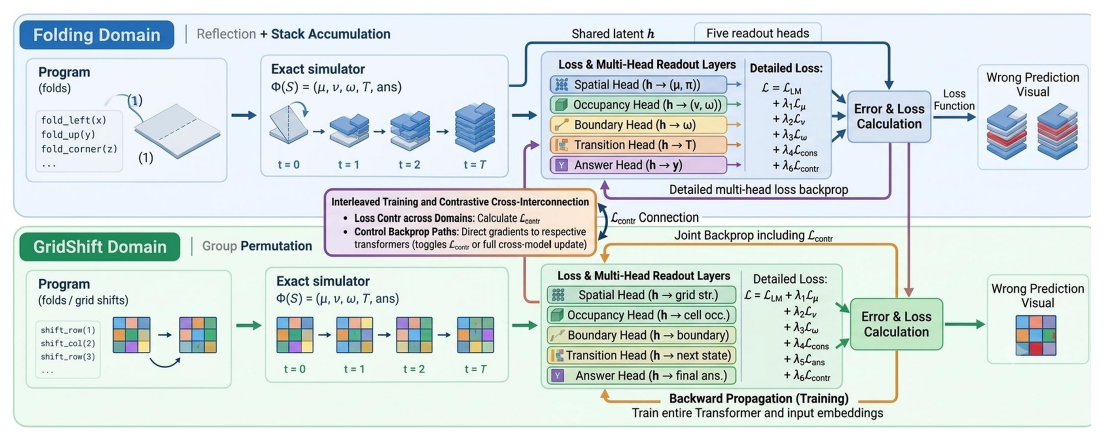}
\caption{The ORION architecture on both domains. An exact simulator emits the canonical state
$\Phi(S) = (\mu, \nu, \omega, T, \mathrm{ans})$ at every execution step; five read-out heads decode
it from one shared latent, and their losses are added to the language-modelling objective. The
answer head is one probe among several rather than the privileged output. The identical wiring runs
on \folding{} and \gridshift{}, which is what lets the contrastive term act across domains.}
\label{fig:architecture}
\end{figure}

Figure~\ref{fig:architecture} shows how the supervision attaches to the backbone. A single shared latent $\mathbf{h}$ feeds five read-out heads --- spatial, occupancy, boundary, transition, and answer --- and the composite loss is the language-modelling term plus the weighted head terms. Two
details matter for interpreting the results. First, the answer is decoded from the same latent as the state, as \emph{one probe among several} rather than as a privileged output; this is what makes the read-out measurements of Figure~\ref{fig:state_tracking} meaningful, since they interrogate the latent directly without autoregressive decoding. Second, the transition head is supervised against
an operator traced from the simulator, so the constraint the model is trained on and the operator the generator uses cannot drift apart (Theorem~\ref{thm:linear}). The same wiring runs unchanged on both domains, which is what the contrastive term across them exploits.

\subsection{Predicted vs Actual Mechanism }
\label{app:mechanism}

\begin{figure}[h]
\centering
\includegraphics[width=0.62\linewidth]{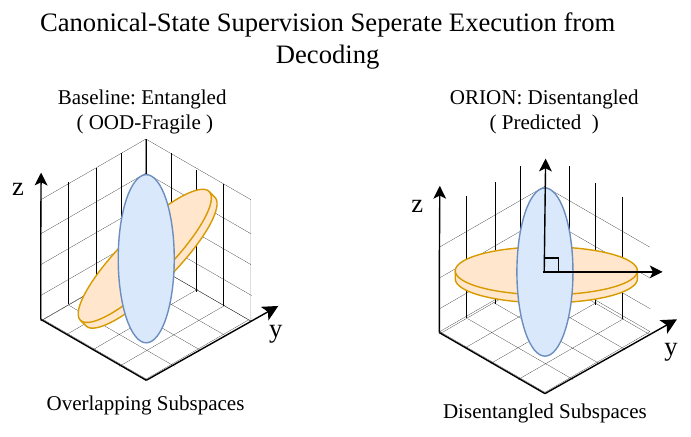}
\caption{The preregistered hypothesis, retained here as the prediction it was. Left: execution and
decoding share overlapping directions in the residual stream, which is the state we expected to make
extrapolation fragile. Right: the predicted effect of canonical-state supervision, separating them
into distinct subspaces. Our diagnostics refute this specific mechanism --- effective dimensionality
rises rather than falls (Table~\ref{tab:geometry}) and the probe subspace is no more causally
necessary than a random one (Appendix~\ref{app:negative}) --- while supporting the weaker claim that
supervision confines the latent trajectory (Figure~\ref{fig:latent_flow}).}
\label{fig:mechanism}
\end{figure}

Figure~\ref{fig:mechanism} states the hypothesis this study preregistered: that entangling execution
with decoding in a single residual stream is what makes a transformer fragile out of distribution,
and that canonical-state supervision would separate the two into distinct subspaces.

The measurements do not support it, and we record that here rather than quietly dropping the
diagram. If supervision disentangled execution from decoding into a lower-dimensional subspace, the
effective dimensionality of the representation should fall; it rises
(Table~\ref{tab:geometry}). If the recovered subspace carried the execution, ablating it should cost
more accuracy than ablating a random subspace of equal rank; it costs the same
(Appendix~\ref{app:negative}). And if the supervised representation were reorganised, cross-model
CKA should drop; it stays between $0.94$ and $0.99$ at every layer.

What survives is a weaker and more specific claim, which is the one the paper makes: supervision
does not rebuild the representation, it changes how the model \emph{moves through} it. The latent
trajectory advances $0.44$ rather than $1.18$ spreads per step along a coherent field
(Figure~\ref{fig:latent_flow}), and the canonical state continues to be decodable past the training
depth when a positional encoding reaches those positions
(Figure~\ref{fig:state_tracking}). Disentanglement was the hypothesis; confinement of the trajectory
is the finding.

\section{The State Algebra}
\label{app:theory}

The theory below is what licenses the auxiliary objectives rather than decoration for them.
Theorem~\ref{thm:cons} says the state is conserved, so a layer map is complete bookkeeping of it;
Theorem~\ref{thm:linear} says the state evolves \emph{linearly} under an operator we can compute
exactly, which is what makes a \emph{label-free} consistency loss possible; Theorem~\ref{thm:equiv}
says syntactically different programs can be provably equivalent, which is what makes a contrastive
loss sound rather than heuristic; and Theorem~\ref{thm:suff} says which statistic deserves the name
\emph{canonical}, which is what control N5 operationalises. All four are numerically verified on
$220$ random programs per domain by the released test suite.

\subsection{Conservation}

\begin{theorem}[Conservation]
\label{thm:cons}
Every fold is a bijection on the element set $E$: each element occupies exactly one stack exactly
once, before and after any fold.
\end{theorem}

\begin{proof}
A fold partitions the footprint into a stationary region and a moved region (straddled diagonal
cells split at quadrant granularity; the quadrant set is closed under the reflections used).
Stationary stacks are copied unchanged; each moved stack is mapped by a reflection $\rho$ --- an
isometry restricted to the grid, hence injective --- and concatenated (reversed) onto its target.
Concatenation neither duplicates nor deletes elements, and injectivity prevents two moved positions
sharing a target.
\end{proof}

\noindent\emph{What this buys.} Nothing is created or destroyed, so the stack \emph{lengths}
$\mu(S)$ are a faithful summary of where material went. That is why $\mu$ is a legitimate
supervision target rather than a lossy proxy.

\subsection{Linear State Transition}

\begin{theorem}[Linear state transition]
\label{thm:linear}
There exist geometry-only index maps $A, B$ with
$\mu(F(S))_i = \mu(S)_{A(i)} + \mu(S)_{B(i)}$; equivalently $\mu(F(S)) = T_F \mu(S)$ for a fixed
$T_F \in \{0,1\}^{|P| \times |P|}$ with at most two ones per row.
\end{theorem}

\begin{proof}
The stack at target $i$ after the fold is the concatenation of the stationary stack previously at
$i$ (or nothing) and the reversal of the stack at the unique moved source $\rho^{-1}(i)$ (or
nothing). Set $A(i) = i$ if stationary else $\varnothing$, and $B(i) = \rho^{-1}(i)$ if in the moved
region else $\varnothing$; both depend only on geometry. Lengths add under concatenation, and a
single reflection gives each target at most one moved source.
\end{proof}

\noindent\emph{What this buys.} Because $T_F$ depends only on the program's geometry and not on the
data, the model can be \emph{constrained} to satisfy $\hat\mu_{t+1} = T_t \hat\mu_t$ without ever
seeing a label for step $t+1$. One supervised first step then pins the entire trajectory. This is
the consistency term of the composite objective, and it is the component that most directly encodes
``execute the program'' rather than ``predict the answer''. The implementation \emph{traces} $A$ and
$B$ through the simulator, so the operator used by the objective cannot drift from the operator used
by the generator.

\subsection{Commutation and Content Equivalence}

\begin{theorem}[Commutation / content equivalence]
\label{thm:equiv}
Axis-orthogonal folds commute on content:
$\sigma(F_H(F_V(S))) = \sigma(F_V(F_H(S)))$; content-equivalent states give identical answers and
layer maps.
\end{theorem}

\begin{proof}
A vertical fold sends $(r, c, q) \mapsto (r, \phi(c), m_V(q))$, independently of $r$, and a
horizontal fold acts only on rows. Both orders send $(r, c, q)$ to
$(\psi(r), \phi(c), m_H m_V(q))$ with $m_H m_V = m_V m_H$, since these are involutions on disjoint
quadrant pairs, so the content maps agree. The answer and layer map are functions of the content map
by Theorem~\ref{thm:suff}, hence coincide.
\end{proof}

\noindent\emph{What this buys.} We can \emph{generate} pairs of distinct programs guaranteed to
share an answer and pull their representations together with a contrastive term, knowing the
supervision is exactly correct rather than assumed. It also gives a free evaluation axis: a model
respecting the algebra should give equivalent programs the same prediction, which is the
\emph{equivalent programs} row of Table~\ref{tab:extra_axes}.

\begin{table}[h]
\centering
\caption{Three further preregistered evaluation axes on the seed set of Table~\ref{tab:main_results}
(mean $\pm$ s.d. over 3 seeds, exact match). \emph{Held-out operation type} withholds diagonal folds
(\textsc{Folding}) and the transpose (\textsc{GridShift}) from training. \emph{Held-out query type}
withholds the multi-cell punch. \emph{Equivalent programs} scores agreement on pairs of syntactically
different programs that Theorem~\ref{thm:equiv} certifies to have the same answer; it is an
in-distribution consistency check, not an extrapolation split.}
\label{tab:extra_axes}
\vspace{0.5em}
\small
\begin{tabular}{lcccc}
\toprule
\textbf{Evaluation axis} & \textbf{Baseline (APE)} & \textbf{RoPE only} & \textbf{ORION only} & \textbf{ORION + RoPE} \\
\midrule
\multicolumn{5}{l}{\cellcolor{orionblue}\textbf{FOLDING}} \\
Held-out operation type & 0.151 $\pm$ 0.012 & 0.150 $\pm$ 0.018 & \textbf{0.175 $\pm$ 0.023} & 0.160 $\pm$ 0.016 \\
Held-out query type & 0.001 $\pm$ 0.001 & \textbf{0.002 $\pm$ 0.002} & 0.000 $\pm$ 0.001 & 0.001 $\pm$ 0.001 \\
Equivalent programs (Thm.~\ref{thm:equiv}) & 0.962 $\pm$ 0.004 & 0.966 $\pm$ 0.002 & 0.966 $\pm$ 0.004 & \textbf{0.973 $\pm$ 0.006} \\
\multicolumn{5}{l}{\cellcolor{orionblue}\textbf{GRIDSHIFT}} \\
Held-out operation type & 0.079 $\pm$ 0.002 & 0.081 $\pm$ 0.003 & 0.081 $\pm$ 0.001 & \textbf{0.085 $\pm$ 0.006} \\
Held-out query type & \textbf{0.000 $\pm$ 0.000} & \textbf{0.000 $\pm$ 0.000} & \textbf{0.000 $\pm$ 0.000} & \textbf{0.000 $\pm$ 0.000} \\
Equivalent programs (Thm.~\ref{thm:equiv}) & \textbf{0.941 $\pm$ 0.010} & 0.929 $\pm$ 0.016 & 0.937 $\pm$ 0.007 & 0.936 $\pm$ 0.036 \\
\bottomrule
\end{tabular}
\end{table}

\subsection{Sufficiency and Identifiability}

\begin{definition}[Cell-content map]
$\tau(S) : (r, c) \mapsto \{\lfloor e/4 \rfloor : e \in \bigcup_q \sigma(S)(r, c, q)\}$ is the set of
original cells present at each footprint position. The \emph{answer map} is
$\mathcal{A}_S : Q \mapsto \mathrm{ans}(S, Q)$.
\end{definition}

\begin{theorem}[Sufficiency and identifiability]
\label{thm:suff}
\textbf{(a)} $\mathrm{ans}(S, Q) = \bigcup_{(r,c) \in Q} \tau(S)(r, c)$, independent of the program.
\textbf{(b)} $\tau(S)$ is recoverable from single-cell answers, so any statistic $\Psi$ sufficient
for the answer map refines $\tau$; hence $\tau$ is the coarsest sufficient statistic, unique up to
relabelling.
\end{theorem}

\begin{proof}
(a) A single punch at $(r, c)$ returns $\tau(S)(r, c)$ by definition, and multi-punch answers are the
union; both reference $S$ only through $\tau$. (b) Recoverability is (a) with $Q = \{(r,c)\}$. If
$\Psi$ is sufficient, $\mathcal{A}_S$ is a function of $\Psi(S)$; composing with recovery, $\tau$ is
a function of $\Psi$. Any statistic coarser than $\tau$ collapses states with different answers, so
is not sufficient.
\end{proof}

\begin{remark}[This is what grounds control N5]
$\tau$ is \emph{minimal sufficient}. ORION's $\Phi$ refines $\tau$ in the geometric coordinates where
the transition is linear (Theorem~\ref{thm:linear}); any bijective relabelling of $\Phi$ is
\emph{equally sufficient} but destroys that structure. Control N5 supervises exactly such a
sufficient-but-non-canonical statistic. If N5 matches ORION, then canonicality is irrelevant and only
exactness mattered --- a clean, preregistered way to lose. On the factorial splits
(Table~\ref{tab:negative_controls}) N5 does match ORION at equal positional encoding, which is why
\S\ref{sec:negative_controls_results} reports the controls as bounding rather than isolating the effect.
\end{remark}

\subsection{The \gridshift{} Analogues}

The same three structural properties hold in \gridshift{}, for entirely different reasons, which is
why it is a fair second test rather than a restatement. Conservation holds because a shift is a
bijection on cells, with no stacking needed. Linear transition holds because the state evolves as
$s_{t+1} = T_t s_t$ with $T_t$ a $0/1$ permutation matrix --- the same operator \emph{form} as $T_F$,
but arising from a group action rather than from reflection and concatenation. Commutation holds
because shifts on disjoint rows or columns commute. The generator emits the identical data schema, so
the model, losses, trainer and evaluator run on it unchanged.

\subsection{A note on positional commutativity}

Since $\mathbf{R}_{\Theta,m}$ acts as a block-diagonal rotation on two-dimensional subspaces, it
preserves inner products of canonical state features under a global index shift:
$\langle \mathbf{R}_{\Theta,m} \Phi(S_t), \mathbf{R}_{\Theta,m+k} \Phi(S_t) \rangle
= \langle \Phi(S_t), \mathbf{R}_{\Theta,k} \Phi(S_t) \rangle$. This is the standard relative-position
property of RoPE. We record it because it motivated the design, but we emphasise that it does
\emph{not} establish that the two mechanisms are orthogonal in any functional sense --- our
measurements show them to be interdependent, since canonical supervision is inert without a
positional encoding that reaches the unseen positions.

\section{Additional Experiments}
\label{app:additional}

This appendix reports the experimental phases that did not fit the main text, including those whose
outcome was neutral or negative. Every table here is generated directly from the released run
artefacts.

\subsection{Additional Evaluation Axes}
\label{app:extra}

Beyond the depth and size axes, the preregistration specified three further splits
(Table~\ref{tab:extra_axes}). Holding out an \emph{operation type} (diagonal folds in \folding{},
the transpose in \gridshift{}) is hard for every condition, and it is the one axis where canonical
supervision without RoPE is the best condition on \folding{} ($0.175$), though the margin lies within
the seed spread. Holding out the \emph{query type} defeats all conditions on both domains: this is a
complete null, and we report it rather than omit it. Agreement on \emph{provably equivalent
programs} --- pairs certified by Theorem~\ref{thm:equiv} to share an answer --- is high for every
condition, so respecting the algebra in-distribution is not what separates the models.



\subsection{Alternative Sequence Architectures}
\label{app:arch}

Any claim that positional structure drives depth extrapolation has to contend with backbones that
carry sequence priors of their own. We therefore trained a state-space model, Hyena, and RWKV at a
matched budget (Table~\ref{tab:architectures}). RWKV is the strongest non-rotary backbone on the
depth axis in both domains, and on \gridshift{} it nearly doubles the APE transformer
($0.175$ against $0.093$), which is consistent with recurrence supplying part of what rotary
embeddings supply. None of the three approaches the range that RoPE reaches, and two of them do not
fit the training distribution well enough for their extrapolation numbers to be interpretable.

\begin{table}[h]
\centering
\caption{Alternative sequence backbones at a matched training budget (6,000 steps, 3 seeds,
mean $\pm$ s.d.). The recurrent and long-convolution models were run because they carry sequence
priors of their own and are therefore confounds for any claim about positional structure. The
state-space model does not fit the training distribution in either domain and Hyena fits only
\textsc{Folding}; RWKV fits both and is the strongest non-rotary backbone on the depth axis
($0.099$ and $0.175$), but none of the three comes near what rotary embeddings reach ($0.651$ and
$0.575$).}
\label{tab:architectures}
\vspace{0.5em}
\small
\begin{tabular}{lccc}
\toprule
\textbf{Backbone} & \textbf{In-domain} & \textbf{Depth ($2\times D$)} & \textbf{Size ($S = 8$)} \\
\midrule
\multicolumn{4}{l}{\cellcolor{orionblue}\textbf{FOLDING}} \\
SSM (S4-style) & 0.349 $\pm$ 0.021 & 0.004 $\pm$ 0.002 & 0.098 $\pm$ 0.009 \\
Hyena & 0.902 $\pm$ 0.003 & 0.022 $\pm$ 0.004 & 0.157 $\pm$ 0.008 \\
RWKV & 0.963 $\pm$ 0.005 & 0.099 $\pm$ 0.012 & 0.160 $\pm$ 0.004 \\
Transformer (APE) & 0.970 $\pm$ 0.003 & 0.021 $\pm$ 0.008 & 0.171 $\pm$ 0.009 \\
ORION + RoPE & 0.985 $\pm$ 0.003 & 0.667 $\pm$ 0.018 & 0.196 $\pm$ 0.010 \\
\midrule
\multicolumn{4}{l}{\cellcolor{orionblue}\textbf{GRIDSHIFT}} \\
SSM (S4-style) & 0.362 $\pm$ 0.012 & 0.034 $\pm$ 0.017 & 0.205 $\pm$ 0.004 \\
Hyena & 0.549 $\pm$ 0.016 & 0.171 $\pm$ 0.005 & 0.197 $\pm$ 0.007 \\
RWKV & 0.835 $\pm$ 0.016 & 0.175 $\pm$ 0.017 & 0.291 $\pm$ 0.009 \\
Transformer (APE) & 0.952 $\pm$ 0.011 & 0.093 $\pm$ 0.012 & 0.314 $\pm$ 0.005 \\
ORION + RoPE & 0.947 $\pm$ 0.018 & 0.598 $\pm$ 0.038 & 0.308 $\pm$ 0.006 \\
\bottomrule
\end{tabular}
\end{table}

\subsection{Robustness to Input Corruption}
\label{app:robust}

Table~\ref{tab:robustness} perturbs the program at inference. Canonical supervision produces
marginally higher exact match at every noise level and an identical score when an operation is
dropped, with differences of $1$--$2$ points that lie within single-seed variation. We take this as a
null result: whatever stabilisation the latent-flow analysis detects (Figure~\ref{fig:latent_flow})
does not surface as robustness in the decoded answer.

\begin{table}[h]
\centering
\caption{Robustness to input corruption on \textsc{Folding} (seed 0, in-distribution test set).
\emph{Token noise} replaces a fraction $p$ of program tokens at random. \emph{Dropped operation}
removes one operation from the program and scores against the answer for the shortened program.
Canonical supervision leaves both curves essentially unchanged, so the robustness it confers is not
visible at the level of the decoded answer.}
\label{tab:robustness}
\vspace{0.5em}
\small
\begin{tabular}{lcccc}
\toprule
& \multicolumn{2}{c}{\textbf{Baseline}} & \multicolumn{2}{c}{\cellcolor{orionblue}\textbf{ORION}} \\
\cmidrule(lr){2-3}\cmidrule(lr){4-5}
\textbf{Corruption} & EM $\uparrow$ & F1$\uparrow$ & EM $\uparrow$& F1$\uparrow$ \\
\midrule
$p = 0.00$ & 0.975 & 0.981 & 0.980 & 0.985 \\
$p = 0.02$ & 0.853 & 0.880 & 0.868 & 0.892 \\
$p = 0.05$ & 0.718 & 0.782 & 0.740 & 0.790 \\
$p = 0.10$ & 0.562 & 0.645 & 0.575 & 0.666 \\
\midrule
\cellcolor{orionblue}Dropped operation ($n = 319$) & \cellcolor{orionblue}0.994 & \cellcolor{orionblue}0.994 & \cellcolor{orionblue}0.994 & \cellcolor{orionblue}0.994 \\
\bottomrule
\end{tabular}
\end{table}

\subsection{Representation Geometry}
\label{app:geometry}

\begin{figure}[h]
\centering
\includegraphics[width=\linewidth]{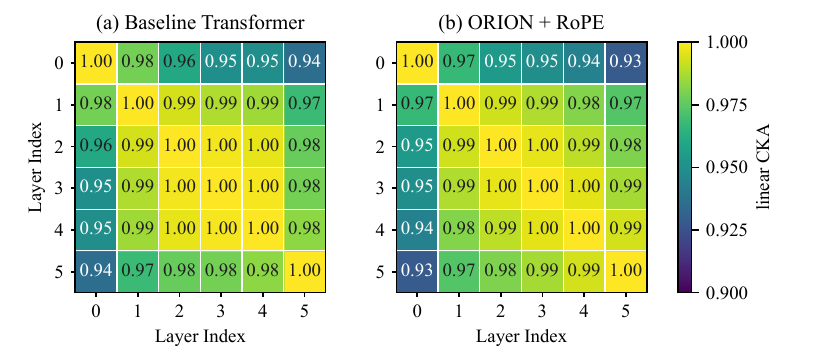}
\caption{Layer-to-layer linear CKA within each model on \folding{} (seed 0, $384$ in-distribution
prompts). Both models show the same smooth block structure, with adjacent layers above $0.97$ and the
embedding layer the most distinct. Canonical supervision does not reorganise the layer hierarchy.
Note the colour scale starts at $0.90$; on an absolute scale the two panels are nearly identical,
which is the point.}
\label{fig:cka}
\end{figure}

We preregistered the hypothesis that canonical supervision compresses the execution subspace into
fewer effective dimensions. Table~\ref{tab:geometry} refutes it. Every variance-based measure moves
in the opposite direction, and layer-wise CKA shows that the supervised model's representation stays
close to the unsupervised model's throughout the stack
(Figure~\ref{fig:cka}). The mechanism is not compression; it is the traversal of the representation,
which is what Figure~\ref{fig:latent_flow} measures.

\begin{table}[h]
\centering
\caption{Representation geometry of the final layer on \textsc{Folding} (seed 0, $n = 600$
in-distribution prompts). The preregistered hypothesis was that canonical supervision compresses the
execution subspace. It does not: the participation ratio, the spectral entropy and the number of
dimensions needed for $90$--$95\%$ of the variance all \emph{rise} under supervision. The TwoNN
estimator is the one measure that moves the other way, so we report the evidence as mixed in
direction but uniform in conclusion --- no measure supports compression. Layer-wise linear CKA
between the supervised and unsupervised models stays high throughout, so the supervised
representation is a nearby one rather than a different one.}
\label{tab:geometry}
\vspace{0.5em}
\small
\begin{tabular}{lcc}
\toprule
\textbf{Measure} & \textbf{Baseline} & \cellcolor{orionblue}\textbf{ORION} \\
\midrule
Participation ratio (of 384 dims) & 10.4 [9.8, 10.6] & 14.3 [13.3, 14.6] \\
TwoNN intrinsic dimension & 12.2 & 10.1 \\
Spectral entropy (nats) & 2.97 & 3.31 \\
Dimensions for 90\% variance & 19 & 26 \\
Dimensions for 95\% variance & 31 & 39 \\
\midrule
Cross-model CKA, \textsc{Folding} & \multicolumn{2}{c}{0.96--0.99 across layers} \\
Cross-model CKA, \textsc{GridShift} & \multicolumn{2}{c}{0.90--0.97 across layers} \\
\bottomrule
\end{tabular}
\end{table}

\subsection{Degradation Curves and the Joint Surface}
\label{app:surface}

\begin{figure}[h]
\centering
\includegraphics[width=\linewidth]{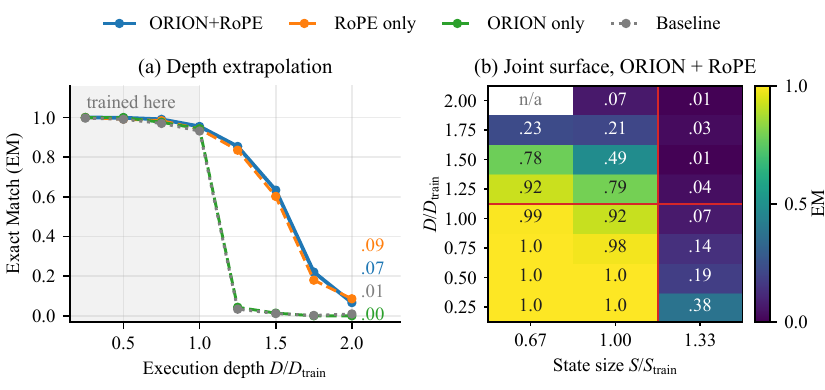}
\caption{Degradation on \folding{}, from the reduced-budget surface sweep. (a) Exact match against
execution depth at the training state sizes; the shaded band is the trained range and the printed
values are the endpoints at $2\times$. The rotary conditions degrade gradually while the two
absolute-position conditions fall off a cliff at the first unseen depth. (b) The joint surface for
ORION + RoPE over both axes, every cell labelled with exact match; red lines mark the edges of the
training region and \emph{n/a} marks a combination the generator cannot produce. Reading up the left
columns shows depth costing accuracy gradually; reading across to the right column shows the size
axis costing all of it at once, at every depth. The size axis is not drawn as a separate curve
because all four conditions coincide on it ($0.195$, $0.190$, $0.169$ and $0.170$ at
$S/S_{\text{train}} = 1.33$).}
\label{fig:surface}
\end{figure}

Figure~\ref{fig:surface} resolves the aggregate numbers of Table~\ref{tab:main_results} into a
degradation curve and a grid over both axes. Three features are worth noting.

First, the two rotary conditions separate from the two absolute ones immediately past the training
boundary, and they separate completely: at $1.25\times$ the training depth, ORION + RoPE and RoPE
hold $0.854$ and $0.835$ while ORION and the baseline have already fallen to $0.043$ and $0.033$.
The failure is a cliff, not a slope, and which side of it a model lands on is decided by the
positional encoding.

Second, between the two rotary conditions the margin is real but small and not uniform: ORION + RoPE
leads RoPE at $1.25\times$, $1.5\times$ and $1.75\times$ ($0.854$ against $0.835$, $0.634$ against
$0.602$, $0.221$ against $0.180$) and trails it at $2\times$ ($0.067$ against $0.086$), where both are
near zero. What canonical supervision buys is a shift in where the cliff falls, not its removal.

Third, the size axis is the horizontal axis of the surface, and its rightmost column is flat in a way
no amount of depth margin repairs: every cell drops to $0.01$--$0.38$ as soon as
$S/S_{\text{train}} = 1.33$, even at a quarter of the training depth. We do not plot the size axis as
a separate curve because all four conditions coincide on it ($0.195$, $0.190$, $0.169$, $0.170$ at
the held-out size), which is consistent with the untrained embedding for that size token rather than
with a graded failure of generalisation.

\subsection{CLRS Per-Algorithm Breakdown}
\label{app:clrs}

\begin{figure}[h]
\centering
\includegraphics[width=\linewidth]{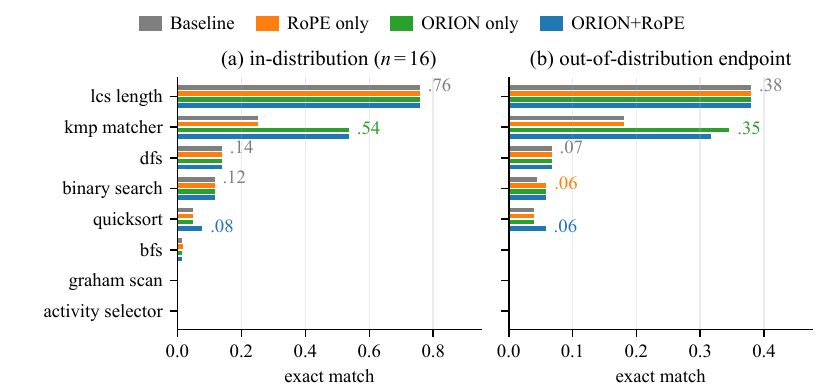}
\caption{Per-algorithm exact match on the CLRS pilot (single seed). (a) In-distribution ($n = 16$).
(b) The preregistered out-of-distribution endpoint, the mean of the length and structural axes. Only
\texttt{kmp\_matcher} separates the conditions; \texttt{activity\_selector} and \texttt{graham\_scan}
are at zero for every condition.}
\label{fig:clrs}
\end{figure}

The aggregate CLRS numbers in Table~\ref{tab:main_results} are dominated by two of the eight
algorithms. Figure~\ref{fig:clrs} separates them. Canonical supervision changes exactly one
algorithm materially --- \texttt{kmp\_matcher}, from $0.25$ to $0.54$ in-distribution and from $0.18$
to $0.35$ out-of-distribution --- while the remaining seven are unchanged to within a point. String
matching is the task in this suite whose hints most closely resemble a canonical execution state,
which makes the result suggestive, but it is one algorithm at one seed and we draw no conclusion from
it.

\subsection{Error Taxonomy}
\label{app:errors}

Table~\ref{tab:errors} classifies the failures. Beyond the training depth, malformed outputs are
rare ($11$ and $3$ of roughly $970$ errors): the models emit well-formed answer sets that are simply
wrong, most often by omitting cells or by being disjoint from the truth. This is the signature of a
state that has drifted rather than of a decoder that has broken, which is the premise the paper is
built on.

\begin{table}[h]
\centering
\caption{Error taxonomy on \textsc{Folding} (seed 0): counts of each failure mode among incorrect
predictions. Beyond the training depth the model almost never emits a malformed or reflected answer:
the failures are well-formed answer sets that miss cells or are disjoint from the truth, which is the
signature of a state that has drifted rather than of a decoder that has broken.}
\label{tab:errors}
\vspace{0.5em}
\small
\begin{tabular}{lcccc}
\toprule
& \multicolumn{2}{c}{\textbf{Baseline}} & \multicolumn{2}{c}{\textbf{ORION}} \\
\cmidrule(lr){2-3}\cmidrule(lr){4-5}
\textbf{Failure mode} & in-domain & depth $2\times D$ & in-domain & depth $2\times D$ \\
\midrule
Final operation not applied & 9 & 92 & 3 & 107 \\
First operation not applied & 0 & 3 & 0 & 5 \\
Query reflected & 2 & 4 & 2 & 5 \\
Answer misses cells & 5 & 369 & 4 & 352 \\
Answer adds cells & 3 & 14 & 1 & 14 \\
Partial overlap & 6 & 195 & 6 & 139 \\
Disjoint from truth & 8 & 282 & 5 & 341 \\
Malformed output & 0 & 11 & 0 & 3 \\
\midrule
\cellcolor{orionblue}\textbf{Total errors} & \cellcolor{orionblue}33 & \cellcolor{orionblue}970 & \cellcolor{orionblue}21 & \cellcolor{orionblue}966 \\
\bottomrule
\end{tabular}
\end{table}

\subsection{Preregistered Diagnostics That Did Not Support the Hypothesis}
\label{app:negative}

Three preregistered analyses returned null or negative results, and we report them in full.

\textbf{Causal necessity of the probe subspace.} Ablating the rank-$8$ subspace recovered by the
canonical-state probe from ORION's final layer changes exact match by $+0.000$ (intact $0.882$),
against $+0.005$ for a random subspace of the same rank and $+0.150$ for the top-$8$ PCA subspace.
The probe subspace is therefore \emph{not} more necessary than a random one by this test, and the
preregistered ``probe $>$ random'' criterion fails.

\textbf{Counterfactual latent editing.} Editing the latent towards the canonical state of a
counterfactual program produced the counterfactual answer in $0$ of $200$ attempts, for both the
supervised and the unsupervised model. We cannot distinguish an absent causal pathway from an
inadequate editing procedure, and we make no claim either way.

\textbf{Statistical power.} With three seeds, the paired seed-level effect on \folding{} depth
extrapolation is $+0.008$ with a between-seed standard deviation of $0.006$ (Cohen's $d = 1.32$).
The minimum detectable effect at $80\%$ power is $0.010$, and five seeds would be required. The
per-example paired tests reported in the main text have far greater power because they pair on the
$1{,}000$ test instances, but the seed-level comparison is underpowered and we do not rely on it.

\section{Experimental Setup and Reproducibility}
\label{app:setup}

\subsection{Dataset Specifications}

\begin{itemize}
    \item \textbf{\folding{}:} Flat-folding of an $n \times n$ sheet at quadrant resolution.
    Training programs apply $1$--$4$ folds to sheets of size $4$ or $6$; the extrapolation splits
    apply $5$--$8$ folds, use size $8$, or both. Diagonal folds are held out for the operation-type
    axis.
    \item \textbf{\gridshift{}:} Cyclic row and column shifts of an $N \times N$ grid of origin
    labels. Training programs apply $1$--$4$ shifts to grids of size $4$ or $6$; the extrapolation
    splits apply $5$--$8$ shifts, use size $8$, or both. Transpose is held out.
    \item \textbf{CLRS pilot:} $8$ algorithms from the CLRS suite in textual form, evaluated at
    problem sizes $16$ (in-distribution), $32$ (length axis) and $64$ (size axis). Negative controls
    were not trained on CLRS.
\end{itemize}

\subsection{Hyperparameters}

\begin{table}[h]
\centering
\caption{Hyperparameter settings across model configurations. The scaling ladder of
Table~\ref{tab:scaling} varies $d_{\text{model}} \in \{256, 384, 512\}$ at a reduced step budget with
five seeds; all other experiments use the configuration below.}
\label{tab:hyperparameters}
\vspace{0.5em}
\small
\begin{tabular}{lcccc}
\toprule
\textbf{Parameter} & \textbf{Baseline} & \textbf{RoPE Only} & \textbf{ORION Only} & \textbf{ORION + RoPE} \\
\midrule
Layers ($L$) & 6 & 6 & 6 & 6 \\
Hidden dim ($d_{\text{model}}$) & 384 & 384 & 384 & 384 \\
Attention heads & 6 & 6 & 6 & 6 \\
Parameters & 11.2M & 11.2M & 11.4M & 11.4M \\
Optimizer & AdamW & AdamW & AdamW & AdamW \\
Learning rate & 0.0003 & 0.0003 & 0.0003 & 0.0003 \\
Weight decay & 0.01 & 0.01 & 0.01 & 0.01 \\
Batch size & 256 & 256 & 256 & 256 \\
Training examples & 40,000 & 40,000 & 40,000 & 40,000 \\
Test instances per split & 1,000 & 1,000 & 1,000 & 1,000 \\
Auxiliary loss weights ($\lambda$) & N/A & N/A & Default$^{*}$ & Default$^{*}$ \\
Total training steps & 6,000 & 6,000 & 6,000 & 6,000 \\
Random seeds & 3 (0--2) & 3 (0--2) & 3 (0--2) & 3 (0--2) \\
\bottomrule
\end{tabular}
\end{table}

\noindent $^{*}\lambda_{\text{spatial}} = \lambda_{\text{topo}} = \lambda_{\text{occ}} =
\lambda_{\text{probe}} = 0.5$, $\lambda_{\text{cons}} = 0.25$, $\lambda_{\text{contr}} = 0.1$. These
were fixed at their preregistered values and never swept; \S\ref{sec:sensitivity_results} varies the
\emph{availability} of supervision, not its weight.

\subsection{Negative Controls in Detail}
\label{app:controls}

Each control keeps ORION's architecture, heads, parameter count and every loss term identical to the
full model, and changes only the content of the supervision target. The output-token targets are
never touched, so the controls degrade the state supervision, not the task. All five were trained
with absolute positional encodings.

\begin{itemize}
    \item \textbf{N1 (Compute-matched):} no auxiliary loss, trained for $1.5\times$ the step budget
    ($9{,}000$ against $6{,}000$ steps). This isolates additional optimisation.
    \item \textbf{N2 (Random targets):} canonical targets replaced by deterministic per-example
    pseudo-random targets of matched dimension --- capacity and a learnable auxiliary gradient, with
    no state content.
    \item \textbf{N3 (Inconsistent state):} occupancy and topology targets corrupted for a fraction
    $\epsilon = 0.25$ of examples, so the state components mutually disagree at a controlled rate.
    \item \textbf{N4 (Shuffled targets):} canonical targets permuted across the batch, so each head
    decodes another example's exact state, $\Phi(S_{\pi(t)})$ --- well-formed but uninformative.
    \item \textbf{N5 (Hashed statistics):} a fixed per-seed coordinate permutation applied to the
    state targets. This is a bijective, information-preserving but \emph{non-canonical} encoding: the
    geometric layout and the linear transition operator are destroyed, so the consistency term is
    disabled for this variant. The answer target is left correct. This is the control that
    Theorem~\ref{thm:suff} and its remark are designed to make decisive.
\end{itemize}

\begin{table}[h]
\centering
\caption{OOD exact match on the depth-extrapolation split ($2 \times D$ at the training state size)
across negative controls, mean $\pm$ s.d. over 3 seeds. CLRS controls were not trained. The controls
lie within the seed spread of the APE baseline and of ORION with APE ($0.026$ on \folding{}, $0.063$
on \gridshift{}); the separation in this table is between the rotary column and everything else.}
\label{tab:controls_depth}
\vspace{0.5em}
\small
\setlength{\tabcolsep}{3.5pt}
\resizebox{\textwidth}{!}{%
\begin{tabular}{lcccccc}
\toprule
\textbf{Benchmark} & \textbf{Baseline (N1)} & \textbf{N2} & \textbf{N3} & \textbf{N4} & \textbf{N5} & \textbf{ORION + RoPE} \\
\midrule
\textbf{FOLDING} & 0.021 $\pm$ 0.008 & 0.021 $\pm$ 0.006 & 0.024 $\pm$ 0.008 & 0.027 $\pm$ 0.008 & 0.031 $\pm$ 0.005 & \cellcolor{orionblue}0.667 $\pm$ 0.018 \\
\textbf{GRIDSHIFT} & 0.093 $\pm$ 0.012 & 0.081 $\pm$ 0.020 & 0.068 $\pm$ 0.018 & 0.077 $\pm$ 0.034 & 0.073 $\pm$ 0.020 & \cellcolor{orionblue}0.598 $\pm$ 0.038 \\
\midrule
\textbf{Mean} & 0.057 & 0.051 & 0.046 & 0.052 & 0.052 & \cellcolor{orionblue}0.633 \\
\bottomrule
\end{tabular}
}
\end{table}

\subsection{Scaling Ladder}
\label{app:scaling}

Table~\ref{tab:scaling} gives the ladder plotted in Figure~\ref{fig:composition}c. The three
widths were trained with five seeds each at a reduced step budget, which is why the absolute
values sit below those of Table~\ref{tab:main_results}; only the within-width comparison between
the two conditions is meaningful.

\begin{table}[h]
\centering
\caption{RoPE versus ORION + RoPE on depth extrapolation in \folding{} across model width. Values
are EM percentages, mean $\pm$ standard deviation over five seeds. The ladder uses a reduced
training budget, so its absolute values are lower than those in Table~\ref{tab:main_results}.}
\label{tab:scaling}
\vspace{0.5em}
\small
\begin{tabular}{lccc}
\toprule
\textbf{Model Scale} & \textbf{RoPE Only Depth OOD} & \textbf{ORION + RoPE Depth OOD} & \textbf{Absolute Delta} \\
\midrule
Small (3.5M) & 41.9 $\pm$ 3.1 & 41.9 $\pm$ 2.7 & +0.0 pp \\
Medium (11.2M) & 54.0 $\pm$ 1.6 & 54.5 $\pm$ 2.3 & +0.4 pp \\
Large (26M) & 54.6 $\pm$ 3.6 & \cellcolor{orionblue}\textbf{62.8 $\pm$ 1.9} & \cellcolor{orionblue}\textbf{+8.2 pp} \\
\bottomrule
\end{tabular}
\end{table}

\subsection{Positional-Encoding Interoperability}
\label{app:pe}

Table~\ref{tab:pe_interoperability} isolates the interaction between positional encoding and state
supervision on \folding{}.

\begin{table}[h]
\centering
\caption{Positional-bias interoperability across the size and depth axes on \folding{}. Values are EM
percentages. ALiBi and other relative-bias schemes were not trained in this study and are therefore
not listed; we make no claim about their interaction with ORION.}
\label{tab:pe_interoperability}
\vspace{0.5em}
\small
\begin{tabular}{lcccc}
\toprule
\textbf{Positional Scheme} & \textbf{State Supervision} & \textbf{Size Extrap.} & \textbf{Depth Extrap.} & \textbf{Joint} \\
\midrule
APE & None & 17.1 & 2.1 & 0.4 \\
APE & ORION & 17.1 & 2.6 & 0.4 \\
RoPE & None & 19.1 & 65.1 & 1.8 \\
\cellcolor{orionblue}\textbf{RoPE} & \cellcolor{orionblue}\textbf{ORION} & \cellcolor{orionblue}\textbf{19.6} & \cellcolor{orionblue}\textbf{66.7} & \cellcolor{orionblue}\textbf{2.2} \\
\bottomrule
\end{tabular}
\end{table}

Read across the rows, the table gives the cleanest statement of the paper's finding. Rotary
embeddings move depth extrapolation from $2.1\%$ to $65.1\%$ with no state supervision at all.
Canonical supervision adds $0.5$ points on top of absolute embeddings and $1.6$ points on top of
rotary embeddings. Neither scheme repairs the size axis.

\subsection{Computational Resources}

\begin{itemize}
    \item \textbf{Compute:} training ran on one node of $4\times$ NVIDIA A100-SXM4-80GB. Summing the
    recorded wall-clock time of every training job in the released manifests gives $40$ GPU-hours
    across the main run, the five-seed follow-up and the architecture baselines. Computing the
    canonical-state targets and their gradients raised per-run wall-clock training time by $181\%$
    relative to the unsupervised baseline. The joint-regime evaluation reported here is inference
    only and was run on CPU from the released checkpoints.
    \item \textbf{Code and artefacts:} implementation code, dataset generators, model weights,
    diagnostic plotting scripts, and the table-generation scripts that emit every table in this paper
    from run artefacts are included in the supplementary material and will be released under the MIT
    License upon publication.
\end{itemize}

\end{document}